\documentclass{article}
\usepackage{ijcai23}

\usepackage{times}
\usepackage{soul}
\usepackage{url}
\usepackage[hidelinks]{hyperref}
\usepackage[utf8]{inputenc}
\usepackage[small]{caption}
\usepackage{graphicx}
\usepackage{amsmath}
\usepackage{amsthm}
\usepackage{booktabs}
\usepackage{algorithm}
\usepackage{algorithmic}
\usepackage[switch]{lineno}

\usepackage{graphicx}
\usepackage{amsmath}
\usepackage{amsthm}
\usepackage{dsfont}
\usepackage{subcaption}
\usepackage{color}
\usepackage{enumerate}
\usepackage{cases}
\usepackage{multicol}
\usepackage{multirow}
\usepackage{enumitem}
\usepackage{bm}
\usepackage{import}
\usepackage{lscape}
\usepackage{amssymb}
\usepackage{xcolor}
\usepackage[algo2e,linesnumbered,ruled,vlined]{algorithm2e}
\usepackage[normalem]{ulem}

\theoremstyle{definition}
\newtheorem{definition}{Definition}[section]

\definecolor{BrightBlue}{RGB}{65, 145, 225}

\newcommand{\armman}{ARMMAN}
\newcommand{\fullname}{Time-series Arm Ranking Index}
\newcommand{\proposed}{TARI}
\newcommand{\numparticipatsfull}{11256}
\newcommand{\numparticipats}{2252}

\makeatletter
\newcommand\footnoteref[1]{\protected@xdef\@thefnmark{\ref{#1}}\@footnotemark}
\makeatother

\title{Limited Resource Allocation in a Non-Markovian World:\\
The Case of Maternal and Child Healthcare}

\author{
Panayiotis Danassis$^1$\and
Shresth Verma$^2$\and
Jackson A. Killian$^1$\and
Aparna Taneja$^2$\And
Milind Tambe$^{1,2}$
\affiliations
$^1$Harvard University\\
$^2$Google Research
\emails
pdanassis@seas.harvard.edu,
vermashresth@google.com,
jkillian@g.harvard.edu,
aparnataneja@google.com,
milind\_tambe@harvard.edu
}

\begin{document}

\maketitle

\begin{abstract}
    The success of many healthcare programs depends on participants' adherence. We consider the problem of scheduling interventions in low resource settings (e.g., placing timely support calls from health workers) to increase adherence and/or engagement. Past works have successfully developed several classes of Restless Multi-armed Bandit (RMAB) based solutions for this problem. Nevertheless, all past RMAB approaches assume that the participants' behaviour follows the Markov property. We demonstrate significant \emph{deviations from the Markov assumption} on real-world data on a maternal health awareness program from our partner NGO, \armman{}. Moreover, we extend RMABs to \emph{continuous state} spaces, a previously understudied area. To tackle the generalised non-Markovian RMAB setting we (i) model each participant's trajectory as a \emph{time-series}, (ii) leverage the power of time-series forecasting models to learn complex patterns and dynamics to predict future states, and (iii) propose the \emph{\fullname{}} (\proposed{}) policy, a novel algorithm that selects the RMAB arms that will benefit the most from an intervention, given our future state predictions. We evaluate our approach on both synthetic data, and a secondary analysis on \emph{real data} from \armman{}, and demonstrate significant increase in engagement compared to the SOTA, deployed Whittle index solution. This translates to $16.3$ hours of additional content listened, $90.8\%$ more engagement drops prevented, and reaching more than twice as many high dropout-risk beneficiaries.
 \end{abstract}

\section{Introduction} \label{sec: Introduction}

According to the latest estimates from 2020~\cite{mmr2020}, the global Maternal Mortality Ratio (MMR) is 152 deaths per 100,000 live births, more than double the UN Sustainable Development Goal (SDG) 3.1's target~\cite{sdg}. For context, the MMR in the USA is estimated to be 35, while in Western Europe is 5. Lack of access to preventive care information, especially in the global south, is a major contributing factor for these deaths. For example, India's MMR is estimated to be 130 deaths per 100,000 live births -- almost 90\% of which are avoidable if women receive the right kind of intervention~\cite{armmanMaternalHealth}. To reduce MMR in India, our partner NGO, \armman{} (\href{https://armman.org/}{armman.org}), employs an automated call-based information program to disseminate critical healthcare information to pregnant women and recent mothers in underserved communities. Such programs have repeatedly demonstrated significant benefits (e.g., see~\cite{HelpMum,mMitra,verma2022saheli,mate2022field,kaur2020effectiveness,pfammatter2016mhealth}), as they raise awareness regarding the need for regular care, potential risk factors, and complications. According to \armman{}, one of the biggest challenges these programs face is that of dwindling adherence, as a large fraction of beneficiaries often drop out. It is thus crucial to provide timely interventions through support calls, or home visits from healthworkers so as to minimize disengagement.

\begin{figure}[t!]
    \centering
    \includegraphics[width=\linewidth, trim={0em 0em 0em 0em}, clip]{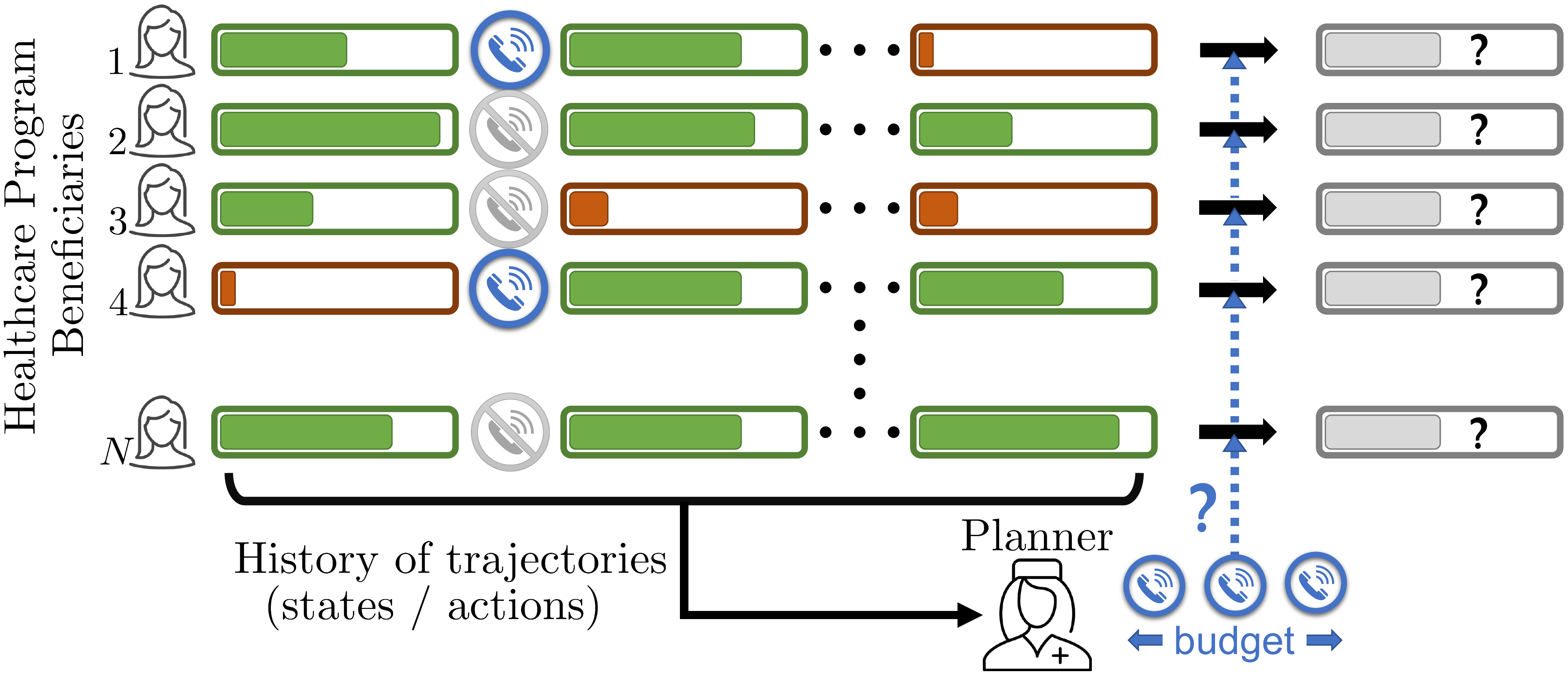}
    \caption{Scheduling healthcare interventions: at each timestep, a planner selects $k$ out of $N$ arms (healthcare program beneficiaries) to schedule an intervention (e.g., a healthcare worker will call or visit). Each bar represents the state (level of engagement) of each arm at each timestep, which can change even when the arm is not pulled. Green (red) bars represent beneficiaries with state above (below) the desired engagement threshold (e.g., for \armman{} the threshold is set to listening to 25\% of the automated message). Intervening on a beneficiary will increase their engagement in expectation (e.g., see beneficiary \#4). The planner observes the states and adjusts its policy to maximize the number of engaging beneficiaries.}
    \label{fig: Setting}
\end{figure}

In this paper, we study the problem of scheduling healthcare interventions under limited healthcare worker resources. We model this resource optimization problem as a Restless Multi-armed Bandit (RMAB) problem~\cite{whittle1988restless}, in which a planner can act on $k$ out of $N$ arms (beneficiaries) at each timestep (Figure \ref{fig: Setting}). Contrary to stochastic bandits~\cite{auer2002finite}, in RMABs each arm has a state, the reward depends on said state, and the state changes even when the arm is not pulled. Past works have developed RMAB-based solutions for several classes of sequential scheduling problems with limited resources. Examples include anti-poaching patrols~\cite{qian2016restless}, machine maintenance~\cite{glazebrook2006some}, online advertising~\cite{meshram2016optimal}, and, healthcare~\cite{wang2023scalable}. Notably, all past RMAB approaches assume that the arms' behaviour follows the Markov property, where state transitions are history independent ~\cite{puterman2014markov}. We challenge this assumption, as human behaviour is likely to contain \emph{temporal dependencies}, i.e., depend on past states, observations, and actions ~\cite{10.1145/2187836.2187919,early2022non,10.1145/1810617.1810658}. Using real-world data on a maternal health awareness program from our partner NGO, \armman{}, we demonstrate significant deviations from the Markov assumption. Specifically, the log-likelihood of observing the historical trajectories increases (up to $23\%$), as we increase the order $h$ of the underlying model (see Section \ref{sec: Non-Markovian Behaviour in Maternal Healthcare Data}).

Even under the Markov assumption, computing an optimal policy for RMABs is PSPACE-hard~\cite{papadimitriou1994complexity}. Instead, state-of-the-art (SOTA) approaches commonly adopt the Whittle index policy~\cite{whittle1988restless}, an approximate solution that estimates the expected future value (Whittle index) of acting on an arm, and then proceeds to act on the top-$k$ arms with the largest value. 

If we want to capture non-Markovian\footnote{By non-Markovian we refer to any Markov process of order 2 or higher. For details please see Section \ref{sec: Problem Formulation}.} behaviors using SOTA Whittle index based approaches, we would run into computation and data limitations. First, there will be a combinatorial explosion of the state space (an $h$-order Markov process can be viewed as a first order Markov process on the expanded state space, where each `super' state consists of $h$ original states, i.e., $s' = \times_{i=1}^h s_i$). Second, since the Whittle index policy requires to know the underlying Markov decision process (MDP) -- which grows exponentially in both (i) the order of the process and (ii) the discretization of the state -- it would need ever larger datasets to calculate the empirical transition probabilities.

\emph{We are the first to cast limited resource optimization problems into the generalised non-Markovian RMAB setting}. In order to provide a \emph{practical}, and \emph{scalable} solution, we take inspiration from the core idea of the Whittle index -- pull arms with the highest expected gains from pulling -- but we drop the cumbersome MDPs. Instead, we opt to \emph{independently} model each participant's trajectory as a \emph{time-series}, leveraging the power of time-series models to learn complex patterns and dynamics to predict future states. Additionally, we develop the \emph{\fullname{}} (\proposed{}) policy, a novel algorithm that selects the arms that will benefit the most from an intervention, given our model's future state predictions.

Finally, as we are no longer limited by the complexity of the Whittle index, in this work, \emph{we extend RMABs to continuous state spaces} -- a previously understudied area -- without the need for the discretization of the state, thus bypassing approximation losses (e.g., see~\cite{sinha2022robustness}). Combining continuous states and non-Markovian transitions offers additional expressiveness that can more accurately capture behavior transitions and patterns, as we showcase in our results.

\begin{figure}[t!]
    \centering
    \includegraphics[width=\linewidth, trim={0em 0em 0em 0em}, clip]{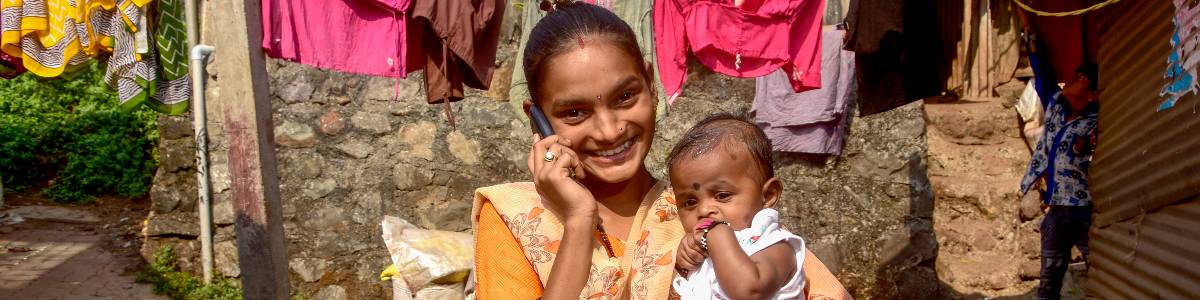}
    \caption{A beneficiary receiving preventive information (photo courtesy of \armman{}).}
    \label{fig: mother}
\end{figure}

\subsection{Our Contributions} \label{sec: Our Contributions}

\noindent
\textbf{(1) We demonstrate significant deviations from the Markov assumption} in real-world data from a \emph{deployed} maternal and child health awareness program by \armman{}.

\medskip
\noindent
\textbf{(2) We are the \emph{first} to cast limited resource optimization problems into the generalised \emph{non-Markovian, continuous state} restless multi-armed bandit setting}, enabling us to capture \emph{temporal dependencies} in human behaviour.

\medskip
\noindent
\textbf{(3) We model each arm as a time-series, and develop a novel algorithm, the \emph{\fullname{}} (\proposed{}) policy}, that acts on arms which will benefit the most from an intervention, given our model's future state predictions, resulting in a \emph{practical}, and \emph{scalable} solution.

\medskip
\noindent
\textbf{(4) We perform a secondary analysis on \emph{real-data} (\numparticipats{} participants, 23 weeks) from a maternal health awareness program (mHealth), in partnership with an Indian NGO, \armman{}}. Compared to the SOTA, deployed Whittle index policy, \proposed{} results in $16.3$ hours of additional content listened, $90.8\%$ more engagement drops prevented, and reaching more than twice as many high dropout-risk beneficiaries.

\subsection{Discussion \& Related Work} \label{sec: Related Work}

\paragraph{Restless multi-armed bandits (RMABs)}

Prior work in RMAB assumes that arms follow the Markov property.
Even in Markovian settings, and when transition dynamics are fully known, RMABs suffer from the curse of dimensionality. Planning an optimal policy is PSPACE-hard~\cite{papadimitriou1994complexity}. As such, SOTA approaches usually deploy approximate planning solutions, most notably the \textbf{\emph{Whittle index policy}}~\cite{whittle1988restless}, which solves the Lagrangian relaxation of the problem. The resulting Lagrange multipliers capture the `resource-efficient value for acting' on an arm (more accurately, the opportunity cost~\cite{Buchanan1991} for not acting). Then the Whittle index policy proceeds to greedily act on the arms with the largest Lagrange multipliers (see Section \ref{supp: Whittle Index Policy} for a formal definition). The Whittle index approach has been shown to be asymptotically optimal (i.e., when $N\rightarrow{}\infty$ with fixed $\frac{k}{N}$)~\cite{weber1990index}, and has been shown to perform well empirically in many applications (e.g.,~\cite{qian2016restless,hsu2018age,mate2022field,kadota2016minimizing}). Nevertheless, the Whittle index remains a heuristic, and asymptotic optimality does not necessarily translate to practically relevant problem sizes and planning horizons, as was recently demonstrated in~\cite{ghosh2022indexability}. Critically, the Whittle index is only optimal under several assumptions (see~\cite{ghosh2022indexability}), which are often hard to validate, and part of active research. Finally, despite being a heuristic, the approach can be prohibitively slow, thus it often requires a problem-specific fast method for computing the index. As such, using the traditional Whittle index on the expanded state space (i.e., $s' = \times_{i=1}^h s_i$) in non-Markov settings is quite challenging and does not guarantee high-quality results, even if the problem can be approximated sufficiently well with low order Markov processes. In this work, we are the first to generalise RMABs to non-Markovian settings. We maintain the key idea of the Whittle index policy (acting on arms that will benefit the most from an intervention), but we use the power of time-series prediction models to capture complex patterns and dynamics to predict the effect of an intervention.

\begin{figure}[t!]
    \centering
    \includegraphics[width=\linewidth, trim={0em 0.9em 0em 0.8em}, clip]{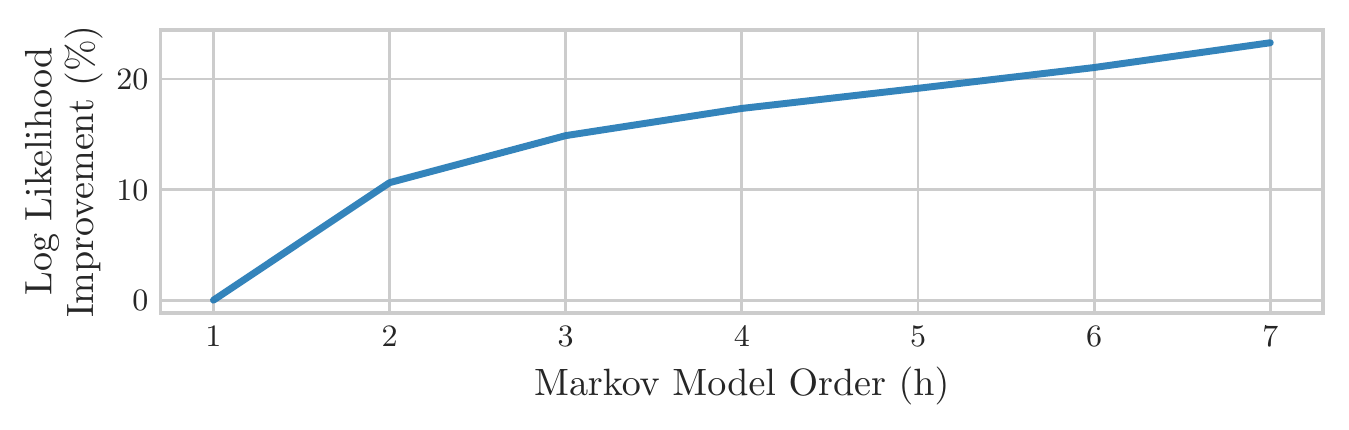}
    \caption{Relative (with respect to $h=1$) improvement in log likelihood. As we increase the order $h$ of the underlying Markov model, the probability of observing the trajectories in our dataset increases. This suggests non-Markovian behaviour.}
    \label{fig: log_likelihood_improvement}
\end{figure}

\paragraph{Time-series Forecasting}

Time series forecasting has been an active area of research over the past few years, with applications in diverse areas such as energy consumption, sensor network monitoring, traffic planning, variations in air pollution, weather forecast, disease propagation, and so on (e.g.,~\cite{matsubara2014funnel,wu2020deep,zhou2021informer}). Solutions range from traditional statistical methods (e.g.,~\cite{ariyo2014arima}), to deep learning-based models (e.g.,~\cite{GuokunLai2017lstm,bai2018empirical,liu2021time}), and, more recently, Transformer-based solutions~\cite{wu2020deep,zhou2021informer,wu2021autoformer,wen2022transformers} which mostly focus on the more challenging long-term forecasting problem (although their effectiveness has recently come into question~\cite{Zeng2022AreTE}). Our contributions are not in developing a SOTA time series prediction model. Instead, we utilise off-the-shelf models in a novel way, to solve non-Markovian RMAB problems. Importantly, our approach is \emph{model agnostic}. In fact, we have evaluated a variety of architectures, including LSTM~\cite{hochreiter1997long}, BiLSTM~\cite{graves2005framewise}, Transformer models~\cite{vaswani2017attention}, adding time-based vector representations~\cite{kazemi2019time2vec}, attention layers, and more (see Section \ref{supp: TSF Model Architecture and Implementation Details}). Future advancements in the area of time series forecasting can easily translate to better performance for the proposed approach.

From the application perspective,~\cite{1525302} and ~\cite{1587138} aim to predict the dropout risk in a similar maternal health awareness setting. Both works are about classification to high and low risk of dropout, and not time-series regression. The former does not optimize or schedule interventions, while the latter aims to identify a smaller subset of beneficiaries and the use the traditional Whittle index in a classic, binary-state Markov setting.

\section{Problem Formulation: Non-Markovian Restless Multi-armed Bandits (NMRMAB)} \label{sec: Problem Formulation}

We consider scheduling problems in which a planner must act on $k$ out of $N$ independent, \emph{continuous} state ($[0, 1]$) arms each round. The planner fully observes the state of each arm, then all arms undergo a history-dependent (i.e., non-Markovian) state transition. The planner's goal is to maximize  the number of processes in `engaging' state over the time horizon $H$.

Let $s_{i, t} \in \mathcal{S}$, and $a_{i,t} \in \mathcal{A}$ denote the state and action taken on arm $i$, respectively, in timestep $t$. We assume that states are continuous in $[0, 1]$, and represent the `level of engagement' of a beneficiary, with higher numbers representing a higher level of engagement. The action set consists of two actions: active ($a_{i,t}=1$), and passive ($a_{i,t}=0$). A non-Markovian Restless Multi-armed Bandit (NMRMAB) problem instance is a 4-tuple $\{\mathcal{N}, k, (X^{i \in N}_t)_{t=1}^H, R\}$, where $\mathcal{N}$ is the set of \emph{independent} arms, $k$ is the budget constraint such that $\sum_i a_{i, t} = k, \forall t$, denoting how many arms can be pulled at a given time-step, $(X^{i \in \mathcal{N}}_t)_{t=1}^H$ is an associated transition process for arm $i$ for time horizon $H$, and $R: (X_t)_{t=1}^H \rightarrow \Re$ is the reward function. In our setting, the reward at timestep $t$ is given by $R(\times_{i < t}(s_i, a_i), s_t) = \mathds{1}_{s_t \geq s^*}$, where $s^*$ is a domain-specific engagement threshold (e.g., \armman{} considers $s^* = 0.25$). The planner's goal is to maximize the total reward, i.e., $\sum_{t \in [1 \dots H]} \sum_{i \in \mathcal{N}} R(\cdot)$. Finally, we assume $(X^{i \in \mathcal{N}}_t)_{t=1}^H$ to be a higher order Markov process. For an $h$-order Markov process, the next state depends on the proceeding $h$ states. More formally:

\begin{definition}[Order $h$ Markov Process]
    Let $x_1, \dots, x_t$ be the elements of the process. A Markov process of order $h$ is a process $(X_t)_{t = 1}^\infty$, such that $\forall t$:
    \begin{align*}
        &\Pr[X_{t} = x_{t} \mid x_{t-1}, x_{t-2}, \ldots,  x_{1}] =\\ &\Pr[X_{t} = x_{t} \mid x_{t-1}, x_{t-2}, \ldots, x_{t-h}]
    \end{align*}
\end{definition}

\section{Non-Markovian Behaviour in Maternal mHealth Data} \label{sec: Non-Markovian Behaviour in Maternal Healthcare Data} 

The proposed modeling raises the question as to whether human activity is indeed non-Markovian in our domain. To answer this question, following related literature~\cite{10.1145/2187836.2187919}, we compute the log-likelihood of the participants' trajectories for a process of order $h$, based on the data on a maternal health awareness program from our partner NGO, \armman{}. Specifically, we start by computing the empirical transition probabilities, assuming the underlying process is of order $h$. Let $h=1$. This is easily achieved by maintaining counters $C_{(s_t, a_t)\rightarrow s_{t+1}}$. The transition probabilities are simply $C_{(s_t, a_t)\rightarrow s_{t+1}} / \sum_{x} C_{(s_t, a_t)\rightarrow x}$. For $h>1$, an $h$-order Markov process can be viewed as a first order Markov process on the expanded state space $s' = \times_{i=1}^h s_i$. Thus, we can employ the same approach to calculate empirical probabilities. Given that we are dealing with continuous states, to maintain said counters, we first need to discretize them. We opted for a binary discretization, as in related literature~\cite{verma2022saheli}, into `engaging' and `non-engaging' states, using $s*$ as a threshold (see also Section \ref{sec: Real Data on Maternal and Child Healthcare}). Then, for each trajectory $x$ in our dataset, we compute $l(h) \triangleq -\log \mathcal{L}(h \mid x) \triangleq -\log \Pr (X=x \mid \text{model of order $h$})$. Finally, we average over all trajectories. Due to the increase in the number of counters and data required, we model up to seventh order processes ($h=7$). Figure \ref{fig: log_likelihood_improvement} shows the relative improvement $-\left(\frac{l(h) - l(h=1)}{l(h=1)}\right)$ in negative log-likelihood for order $h$ Markov processes ($x$-axis). There is a clear improvement for higher order models, specifically about  $10\%$ for $h=2$, and up to $23\%$ for $h=7$. This suggests that, under the often-used binary-discretization model, participants' behavior across the duration of the program is indeed not Markovian.

\section{Methodology} \label{sec: Methodology}

\subsection{RMABs as a Time-series Forecasting Problem} \label{sec: RMABs as a Time-series Forecasting Problem}

In this work, we are the \emph{first} to propose a time-series forecasting (TSF) based framework for supervised representation learning of arms' trajectories in non-Markovian RMABs.

\subsubsection{Preliminaries: TSF Problem Formulation} \label{sec: Preliminaries: TSF Problem Formulation}

Let $\mathcal{X}=\{X_t^{1}, \dots, X_t^{M}\}{_{t=1}^{h}}$ denote the historical data of an $M$-variate series, where $h$ is the look-back window length and  $X_t^{i}$ is the value of the $i$th variate at timestep $t$. Formally, the TSF task is, given $\mathcal{X}$, to predict the future $T$ values, i.e., $\hat{\mathcal{Y}}=\{\hat{Y}_t^{1}, \dots, \hat{Y}_t^{M}\}{_{t=h+1}^{h+T}}$. There are two methods for predicting $\hat{\mathcal{Y}}$ when $T>1$: Iterated multi-step (IMS) forecasting~\cite{taieb2012recursive} where the model learns to predict a single-step forward, and then is recursively called to obtain multi-step forecasts. Alternatively, with direct multi-step (DMS) forecasting~\cite{chevillon2007direct}, one trains a model that directly optimizes the multi-step forecasting objective, and is only called once. Usually IMS forecasts result in smaller variance than DMS, but the error could accumulate over long prediction horizons~\cite{Zeng2022AreTE}. For simplicity, and given the relatively short forecasting horizon in our domain, we opted for the IMS approach (see also Section \ref{sec: Myopic Whittle}).

\subsubsection{Labeled Dataset Using a Sliding Window} \label{sec: Labeled Dataset Using a Sliding Window}

\begin{figure}[t!]
    \centering
    \includegraphics[width=1\linewidth, trim={0em 0em 0em 0em}, clip]{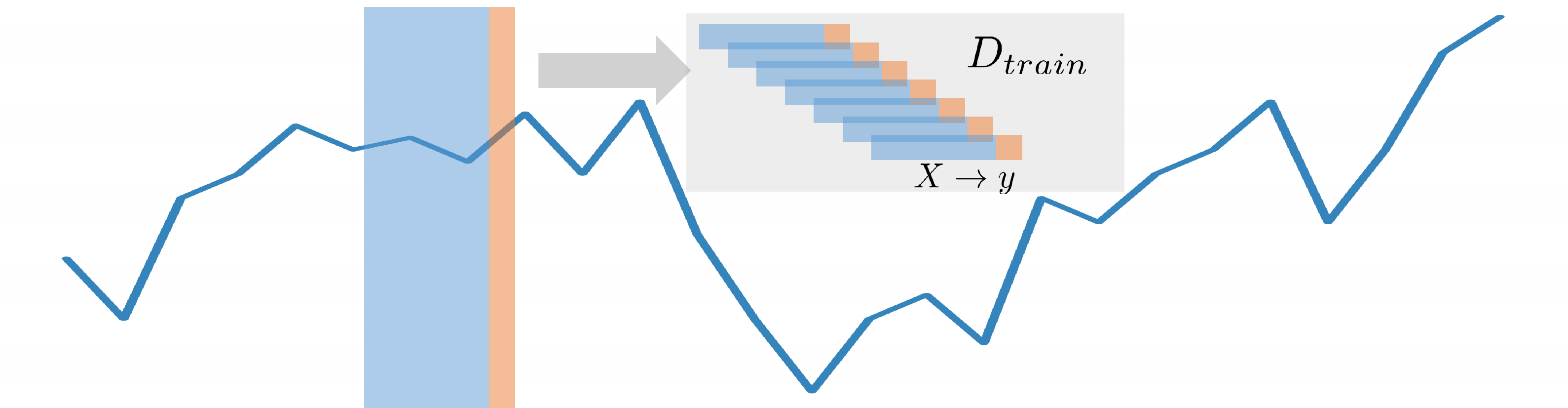}
    \caption{We use a fixed length sliding window, sliding right one time-step at a time, to generate a supervised learning training dataset from participants' time series trajectories.}
    \label{fig: sliding_window}
\end{figure}

To produce a labeled dataset for both training ($D_{train}$), and evaluation ($D_{test}$), we used a fixed-length sliding window approach. First, we assume that the planner has access to an offline historical dataset of beneficiaries' trajectories, as is the case for example with our partner NGO, \armman{}. We then run a sliding window of length $h$ on each trajectory in the dataset to get training samples of history ($\mathcal{X}$), and next state ($\hat{\mathcal{Y}}$), as depicted in Figure \ref{fig: sliding_window}. For the application at hand, $\mathcal{X}=\{s_t, a_t\}{_{t=1}^{h}}$, i.e., pairs of state/actions. This intuitively corresponds to an $h$-order Markov model approximation. $h$ is a hyper-parameter that depends on the application and needs to be tuned depending on (i) the level of non-Markovian behaviour of beneficiaries, and (ii) the achieved error of the model (larger $h$ does not necessarily translate to lower error). Finally, $\hat{\mathcal{Y}}=\{s_{t}\}_{t=h+1}$, i.e., we only predict the next state. Optionally, depending on the domain, we can enhance the input with auxiliary tokens relevant to the task and the arms' behavior (e.g., socio-demographic features).

\subsection{Proposed Approach: \fullname{} (\proposed{}) Policy}  \label{sec: Myopic Whittle}

The proposed \fullname{} (\proposed{}) maintains the core idea of the traditional Whittle index: estimate the expected future value of acting on an arm, and then greedily act on arms that will benefit the most from an intervention. However, we adjust the methodology to account for (i) the additional complexity due to the non-Markovian setting, and (ii) the scarcity of resources. Specifically for the latter, a key challenge in real-world applications (and especially ones related to healthcare), is that they are resource constrained. For example, despite the scale of \armman{}'s operation, with \emph{millions} of active users, due to limited availability of healthworkers beneficiaries typically receive \emph{at most one intervention} (phone call by a health worker) within a period of 3 months. Taking this into account, \emph{\proposed{} estimates the marginal long-term improvement in engagement if you act once on a arm (and never act again), compared to never acting}.

Specifically, we model each arm independently as a time series, and train a model to predict the next state ($s_{t+1}$), given (i) a history of state/actions ($\times_{i<t}^h(s_i, a_i)$) of length $h$, (ii) the current state ($s_t$), and (iii) the potential action ($a_t$, where $a_t=1$ corresponds to acting, and $a_t=0$ not acting). This is the offline training part. We use this model online, in an iterated multi-step manner (see Section \ref{sec: RMABs as a Time-series Forecasting Problem}), to generate a long term forecast ($s_{t+1}, s_{t+2}, \dots, s_{t+H}$), which then we use to compute the \proposed{} index for planning as follows. 

For each arm $n$ \emph{independently}, we estimate two quantities by recursively using our TSF model: (i) The time $u_{n}$ until arm $n$ switches to non-engaging,\footnote{Non-engaging means that the continuous state $s$ drops below an application-specific threshold ($s^*$). \armman{} considers $s^* = 0.25$.} if we act once at timestep $t$ and never act again (line \ref{algo: act} in Algorithm \ref{algo: mWhittle framework}, and green box in Figure \ref{fig: proposed approach}), and (ii) the time $v_{n}$ until arm $n$ switches to non-engaging, if we never act (line \ref{algo: not act} in Algorithm \ref{algo: mWhittle framework}, and orange box in Figure \ref{fig: proposed approach}). Then the \proposed{} index for arm $n$ is simply given by the ratio of the two numbers (Equation \ref{Eq: mWhittle}). This intuitively gives the `value' of acting. Finally, at each timestep, just like with the traditional Whittle index, we act on the top-$k$ arms with the highest \proposed{} value. The proposed approach is depicted in Figure \ref{fig: proposed approach} and Algorithm \ref{algo: mWhittle framework}.
\begin{equation} \label{Eq: mWhittle}
    \text{\proposed{}}(n) = \frac{u_{n}}{v_{n}}
\end{equation}

The \proposed{} policy offers significant advantages. Arms are modeled independently, which allows for scalability. Furthermore, it is computationally efficient to train and compute in non-Markovian continuous state settings, contrary to the traditional Whittle index which requires supporting and computing over MDPs that grow exponentially in both (i) the order of the underlying Markov process, and (ii) the discretization of the state. 

\begin{algorithm2e}[t!]
    \SetKwInput{KwResult}{Offline}
    \KwData{Historical dataset of beneficiaries' trajectories}
    \KwResult{Train TSF model $\mathcal{M}$ to predict the next state}
    \SetKwInput{KwResult}{Online}
    \KwResult{Decision timestep $t$:}
    {
        \For{$arm \in \mathcal{N}$} {
            $a$ \textbf{ = 1}, $u = 1$, $s = s_t$, history=$\times_{i<t}^h(s_i, a_i)$\\
            $s' = \mathcal{M}(\text{history}, s, a)$\\
            \While(\tcp*[h]{While the state is above the engagement threshold. Forecast ahead at most until the time horizon.}){$s' \geq s^*$ and $u \leq H$} { \label{algo: act}
                history.append($(s, a)$)\\
                $s = s'$, $a$\textbf{ = 0}, $u = u + 1$\\
                $s' = \mathcal{M}(\text{history}, s, a)$\\
            }
            
            \textcolor{white}{.} \\
            $a$ \textbf{ = 0}, $v = 1$, $s = s_t$, history=$\times_{i<t}^h(s_i, a_i)$\\
            $s' = \mathcal{M}(\text{history}, s, a)$\\
            \While{$s' \geq s^*$ and $v \leq H$} { \label{algo: not act}
                history.append($(s, a)$)\\
                $s = s'$, $a$\textbf{ = 0}, $v = v + 1$\\
                $s' = \mathcal{M}(\text{history}, s, a)$\\
            }
            
            \textcolor{white}{.} \\
            \proposed{}($arm$) = $\frac{u}{v}$
        }
    }    
    \caption{\fullname{}}
    \label{algo: mWhittle framework}
\end{algorithm2e}

\section{Simulation Setup} \label{sec: Simulation Setup}

Training data are constructed in the manner described in Section \ref{sec: RMABs as a Time-series Forecasting Problem}. We use a $64\%$, $16\%$, $20\%$ split for the training, validation, and testing datasets, respectively. All experiments are averaged over $10$ independent runs.

\subsection{Baselines} \label{sec: Baselines}

We compare the proposed \proposed{} to four baselines: (i) the \textbf{Whittle index policy}~\cite{whittle1988restless}, (ii) \textbf{round-robin} selection, which often corresponds to the default policy for many NGOs~\cite{mate2022field} including \armman{}, (iii) \textbf{random}, where we act on arms selected uniformly at random, and (iv) \textbf{control}, where there are no support calls (no intervention, i.e., $a_{i, t} = 0, \forall i \in \mathcal{N}, \forall t \in H$).

We chose to compare to the Whittle index, as it is a popular, SOTA approach that has been deployed in the real-world (e.g., see~\cite{mate2022field,verma2022saheli}). Informally, the Whittle index of an arm captures the added value from pulling said arm. Consider a `passive subsidy' -- a hypothetical exogenous compensation $m$ rewarded for not pulling ($a=0$) arm $i$. The Whittle index is defined as the smallest subsidy necessary to make the planner indifferent between pulling and not pulling (assuming indexability~\cite{whittle1988restless}), i.e., 
\begin{equation} \label{Eq: Whittle action-indifference}
    W_i(s) \triangleq \inf\nolimits_{m} \{ Q_{i}^m(s \mid a=0) = Q_{i}^m(s \mid a=1)\}
\end{equation}
The Whittle index policy computes the $W_i(s)$ of all arms and pulls the arms with the highest values of the index at each timestep. The augmented (with subsidy $m$) Bellman equations are solved via value iteration, and binary search is used to find the smallest $m$ that satisfies Equation~\ref{Eq: Whittle action-indifference}. To use the Whittle index in our setting, we must first discretize the continuous state. Following the convention in previous deployments (e.g.,~\cite{mate2022field}), we assume a binary state Markov model (`engaging' = 1, `non-engaging' = 0, thresholded at $s^*$). For completeness, we also run simulations with a more fine-grained discretization of the state, and also incorporating history by using an expanded state space (i.e., $s' = \times_{i=1}^h s_i$). Of course this significantly increases computational and memory complexity, and data requirements (see Sections \ref{sec: Introduction}, \ref{sec: Related Work}, \ref{sec: Non-Markovian Behaviour in Maternal Healthcare Data}).

\begin{figure}[t!]
    \centering
    \includegraphics[width=\linewidth, trim={0em 0em 0em 0em}, clip]{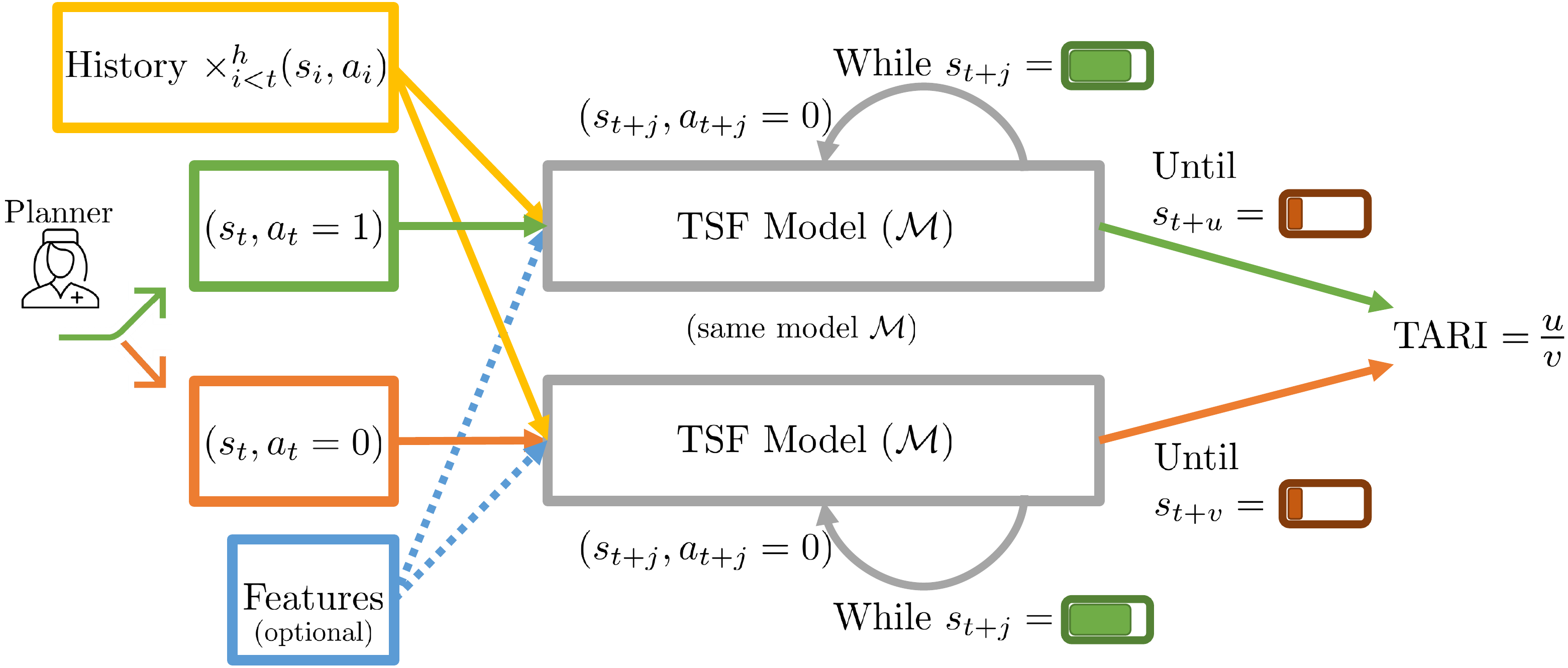}
    \caption{Graphical representation of the calculation of the proposed \proposed{}. We consider two options for the planner: (i) act once (green box, $a_t = 1$) and then never act ($a_{t+j} = 0$), or (ii) never act (orange box). We recursively call the TSF model to predict the next state in the trajectory (feedback loop on the left), until the state switches from engaging (green bar) to non-engaging (red bar). Let this be at timestep $u$ (i.e., $s_{t + u} < s^*$) and $v$ (i.e., $s_{t + v} < s^*$) for options (i), and (ii), respectively. The ratio between the timesteps needed for the switch if we had acted, compared to not, constitutes the \proposed{}. This is computed independently for each arm.}
    \label{fig: proposed approach}
\end{figure}

\subsection{Synthetic Data}  \label{sec: Synthetic Data}

We generate trajectories containing an equal number of the following types of `agents' (arms). (i) \textbf{Habit former}: The value of the continuous sate drops under passive action ($a=0$), and increases with $a=1$. If the state reaches 1 (formed a habit), it stays there for some duration, independent of the action. (ii) \textbf{Motivation based}: State drops over time. If we act, the state returns to baseline. (iii) \textbf{Random}: Random state independent of the action.

Drop rates, increase rates, habit duration, and state baselines include agent specific noise, drawn uniformly at random (UaR). Trajectories used for testing have \emph{higher noise} than the ones used for training (see Section \ref{supp: Synthetic Data}). Historical trajectories are produced by simulating the participants under various simple intervention policies. Specifically, we act $i$ times (drawn UaR in $[6, 24]$), every $j$ timesteps (also drawn UaR in $[1, 14]$). Finally, every participant is associated with a \emph{noisy} `demographic' feature related to their type. These features are given as an additional input to our model (see Figure \ref{fig: proposed approach}). Also, they are used by the Whittle index baseline, which learns (offline) empirical transition probabilities for each type of agent, and then uses the features online to map each arm to the corresponding probabilities.

\subsection{Real Data on Maternal and Child Healthcare}  \label{sec: Real Data on Maternal and Child Healthcare}

We use data from a large-scale maternal and child healthcare program operated by our partner NGO, \armman{}. The program serves pregnant women and early mothers in disadvantaged communities with median daily family income of \$3.22
-- below the global poverty line~\cite{TheWorldBank} -- by disseminating timely health information (via \emph{automated} voice calls) to reduce maternal, neonatal, and child mortality and morbidity. The main challenge the program faces is drop in engagement over time. Engagement is measured in terms of total number of automated voice messages listened. To mitigate this problem, a planner schedules support calls by \emph{limited} healthcare workers.

We model this setting as a continuous state, fully observable RMAB problem. The state of each beneficiary represents the listening time. Each automated voice message has a maximum length of 120 seconds, which we normalise to $[0,1]$. 
The planner's task is to recommend a subset of beneficiaries every week to receive support calls from healthcare workers, with the goal to maximize the number of beneficiaries above the engaging threshold $s^*$. \armman{} considers a beneficiary to be engaging if they listen to more than 30 seconds of the automated message (i.e., if $s>0.25=s^*$). Transition dynamics are unknown, and we make no Markov assumptions. Finally, the dataset also includes socio-demographic features per beneficiary such as age, gestational age, family income, education, etc., that may be used as auxiliary information.

\subsubsection{Training Dataset}  \label{sec: Training Dataset}

We use historical data from a large-scale quality improvement study performed by \armman{} in 2022, obtained with beneficiary consent. The data follows 12000 participants (\numparticipatsfull{} with complete state information by the end) over a period of 31 weeks. In the study, a set of beneficiaries received interventions from a variety of policies (see Section \ref{supp: Real-World Dataset}). Each beneficiary is represented by a single trajectory of states (engagement behavior) and actions (received, or not a call from a healthworker). Demographic features are used to infer the missing transition dynamics for the Whittle index baseline, as in~\cite{verma2022saheli}.

\subsubsection{Notice on Data Usage}  \label{sec: Notice on Data Usage}

Our simulations are a secondary analysis on different evaluation metrics. All data are anonymized, and we have received approval from \armman{}'s ethics board. There is no actual deployment of \proposed{} at \armman{}.

\subsection{Time Series Forecasting Models} \label{sec: Time Series Forecasting Models}

We implemented a variety of time series forecasting models, ranging from simple LSTM, and BiLSTM architectures, to adding time-based vector representations and attention layers, to Transformer models. The majority of the models showed high performance. As such, we opted to use an LSTM-based architecture, as simpler models -- being less computationally intensive and more sustainable in the long run -- would be preferred by NGOs which are operating in a low resource environment. Our chosen model achieved MAE of $0.03$ on average for one step prediction on synthetic data (excluding random agents), and $0.20$ on real data. We used a history of $h=7$ timesteps as input for the synthetic data, and $h=8$ timesteps for the real data evaluation. For detailed results, including long horizon forecasts, please see Section \ref{supp: TSF Model Architecture and Implementation Details}. It is important to note that (i) as discussed in Section \ref{sec: Related Work}, our contribution is not in developing or improving SOTA TSF models, and (ii) the proposed \proposed{} depends on relative trends (ratio of two predicted trajectories, see Equation \ref{Eq: mWhittle}), thus if the model consistently over- or under-predicts the two trajectories, the trend (ratio) will be consistent.

\subsubsection{Counterfactuals in Non-Markovian Settings} \label{sec: Counterfactuals in Non-Markovian Settings}

Under no Markovian assumptions, we can not build a beneficiaries' model to compute counterfactual trajectories for the evaluation on real data. To compute such counterfactuals, we employed a \emph{separate} TSF model trained on the \emph{entire} dataset (train, validation, and test data). This model is \emph{only used when trajectories deviate} from the historical data. Given that we only act on a small percentage of the beneficiaries (about $2\%$), the vast majority of the trajectories follow the historical real data, thus the model will not be used. Note that this second model is only needed for the purposes of the simulation. This is a fully observable RMAB, thus in real life we would actually observe the next state of each participant. We have evaluated an alternative approach to computing counterfactuals (see Section \ref{supp: Counterfactuals in Non-Markovian Settings Alternative}) in the appendix (see Section \ref{supp: Detailed Evaluation Results: Real Data}).

\section{Results on Synthetic Data} \label{sec: Results on Synthetic Data} 

\paragraph{Engagement}

\begin{figure}[t!]
    \centering
    \includegraphics[width=1\linewidth, trim={0em 0.6em 0em 0.7em}, clip]{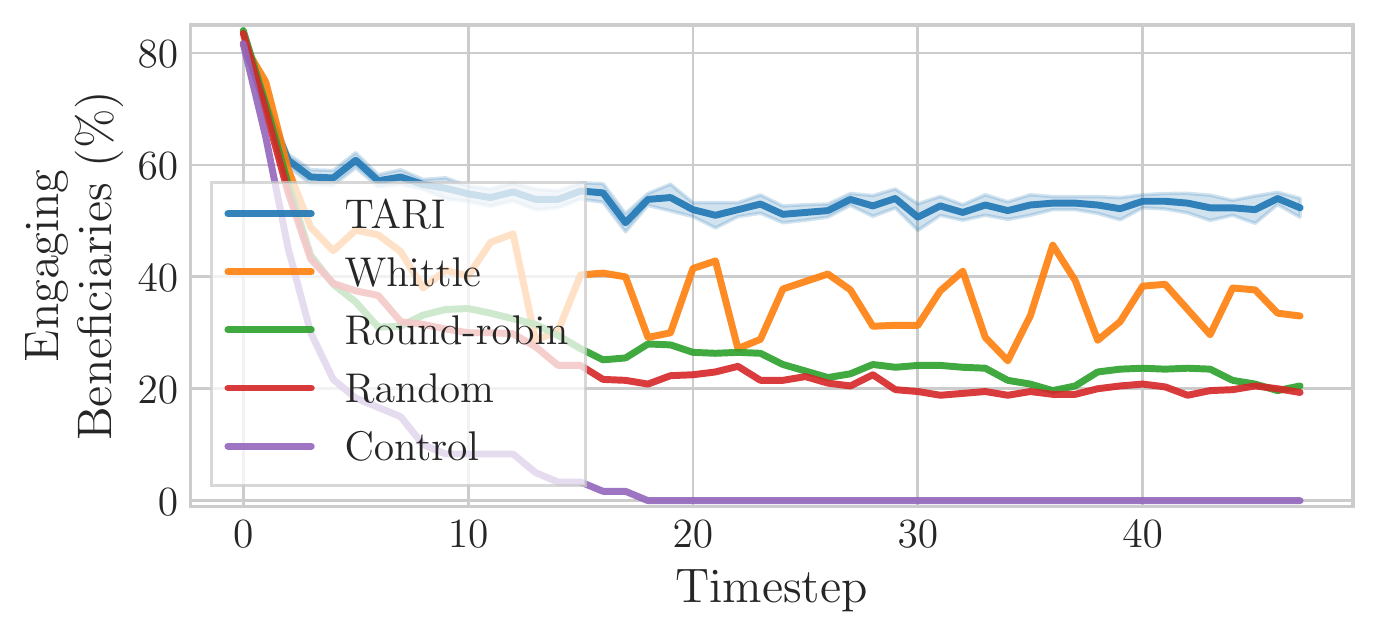}
    \caption{Synthetic data, $N = 90$ arms, budget $k = 9$. Percentage of engaging beneficiaries (excluding the random agents). The proposed approach achieves $44.2\%$ higher engagement on average, and up to $107.3\%$, compared to Whittle index. Note that we opted not to show standard deviations for the baselines for better visualization, due to the high values (contrary, \proposed{} achieves low s.d. of $3.1\%$).}
    \label{fig: results synthetic data}
\end{figure}

We run a long-term simulation of one year (52 timesteps). Starting with the engagement, Figure \ref{fig: results synthetic data} shows the percentage of engaging beneficiaries during the time horizon. \proposed{} achieves $44.2\%$ higher engagement on average (across timesteps and independent runs), and up to $107.3\%$, compared to the Whittle index policy (best baseline). Compared to Round-robin and Random it achieves $102.5\%$ and $128.6\%$ higher engagement on average, respectively. Additionally, \proposed{} achieves significantly lower standard deviation (3.1\%), compared to over $100\%$ for all other baselines, as \proposed{} never intervenes on Random agents.

\paragraph{Robustness}

We evaluated the proposed approach under varying number of arms ($N=\{30, 90, 120, 600\}$) and budgets ($k=\{0.01, 0.1, 0.2\} \times N$, for $n = 90$). In both cases, \proposed{} significantly outperformed all the baselines. Specifically, compared to the Whittle index (best baseline), \proposed{} achieved on average $44.2 - 48.2 \%$ higher engagement when varying the number of arms, and $37.8 - 88.7 \%$ when varying the budget (the smallest budget corresponds to just one intervention, hence the lower improvement). Furthermore, we enhanced the model for the Whittle index baseline to incorporate history ($s' = \times_{i=1}^h s_i$, for up to 4 past states), and a more fine-grained discretization of the state (up to 9 bins) -- recall that the Whittle index baseline is not designed for continuous states, and thus requires discretization. In all of the cases, \proposed{} achieved at least $44.2\%$ higher engagement on average (and up to $144.2\%$). For detailed results, please see Section \ref{supp: Detailed Evaluation Results: Synthetic Data}.

\section{Results on Real Data on Maternal and Child Healthcare} \label{sec: Results on Real Data}

We run our analysis for more than 5 months (31 weeks of data, minus the 8 weeks we give as input to the TSF model, which corresponds to the observation period of \armman{}). Due to limited resources, \armman{} is able to provide support calls to about $2-4\%$ of beneficiaries at each week (timestep). In our dataset of \numparticipats{} beneficiaries (in the test set), the lower number corresponds to just $46$ support calls per week. All policies were evaluated in the entire test set.

\begin{figure}[t!]
  \centering
  \begin{subfigure}[t]{\linewidth}
    \centering
    \includegraphics[width=1\linewidth, trim={0em 0.8em 0em 0.4em}, clip]{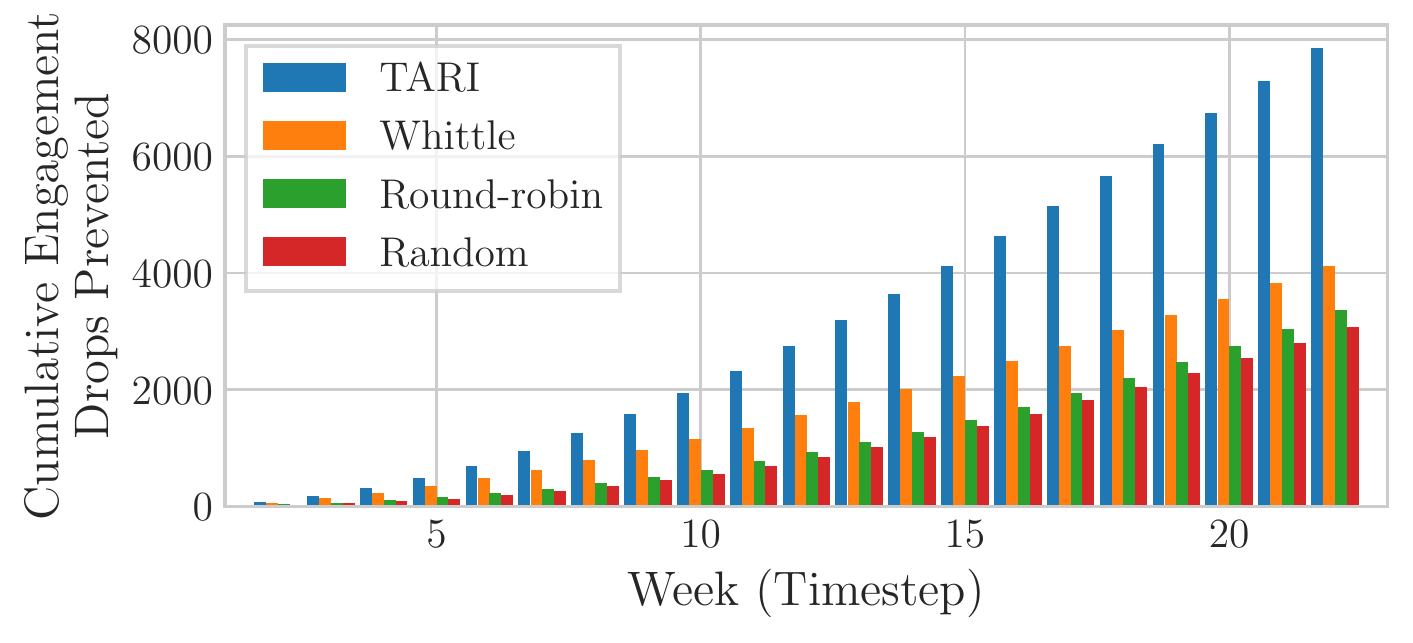}
    \caption{Cumulative engagement drops prevented, compared to no intervention. \proposed{} achieves an \emph{$90.8\%$ improvement over Whittle}.}
    \label{fig: drop prevention}
  \end{subfigure}
  ~ 
  \begin{subfigure}[t]{\linewidth}
    \centering
    \includegraphics[width=1\linewidth, trim={0em 0.8em 0em 0em}, clip]{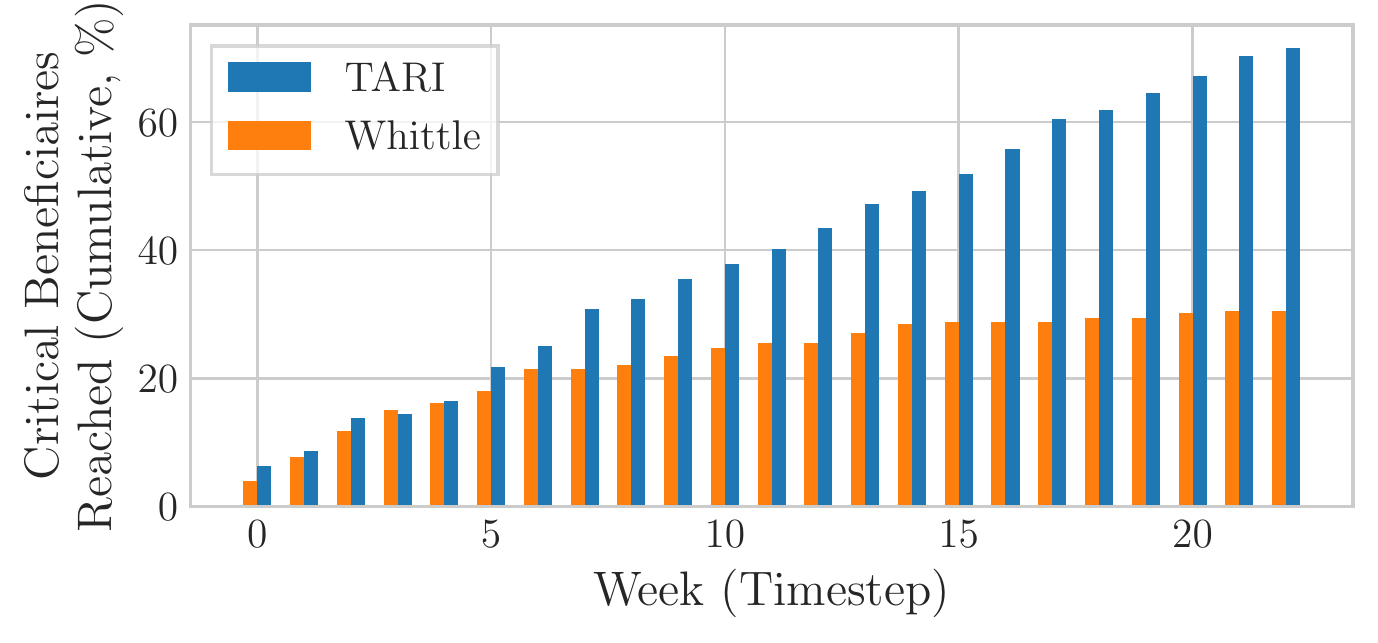}
    \caption{Cumulative percentage of high dropout-risk beneficiaries reached. \proposed{} achieves a \emph{$135.2\%$ improvement over Whittle}.
    }
    \label{fig: critical beneficiaries reached}
  \end{subfigure}
  \caption{Real data on maternal and child healthcare awareness from \armman{}. $N = \numparticipats{}$ beneficiaries (arms), budget $k = 0.02 \times N = 45$. Results are averaged over 10 independent runs.}
\end{figure}

Figure \ref{fig: drop prevention} depicts the cumulative engagement drops ($s<s^*=0.25$) prevented, compared to control. This figure clearly demonstrates the importance, and difficulty of scheduling effective support calls: the naive \emph{round-robin approach (often used by NGOs~\cite{mate2022field}) performs similarly to random}, i.e., following such a policy would incur all the cost associated with providing support calls, without any benefit. \proposed{} achieves an \emph{$90.8\%$ improvement over Whittle}, and $133.9\%$ over round-robin. Ensuring beneficiaries remain \emph{consistently} engaged is crucial for the success of any healthcare program.

\paragraph{Real-world Significance}

To put the real-world significance of \proposed{}'s engagement gains into concrete numbers, this corresponds to $3736$ and $7851$ additional messages listened compared to the Whittle index, and Control, respectively. Or, in other words, \emph{$16.3$ hours of additional content listened by the beneficiaries compared to the Whittle index, and $62.6$ additional hours compared to Control}. Adhering to the program improves health literacy, which would ultimately lead to better health outcomes.

Finally, an important open problem for \armman{} is identifying and proactively reaching beneficiaries with high risk of dropping out from the program (`critical beneficiaries', see Section \ref{supp: High Dropout-Risk Beneficiaries}). Figure \ref{fig: critical beneficiaries reached} shows the cumulative percentage of critical beneficiaries that \proposed{}, and Whittle chose to intervene. By the end of our observation period, \proposed{} had reached \emph{$71.6\%$ of critical beneficiaries}, while Whittle only $30.4\%$, a \emph{$135.2\%$ improvement}. Please see Section \ref{supp: Detailed Evaluation Results: Real Data} for results on engagement, varying budgets, and more.

\section{Conclusion} \label{sec: Conclusion}

In this work, we study for the \emph{first} time non-Markovian Restless Multi-armed Bandits (NMRMAB). Using real-world data on a maternal health awareness program from our partner NGO, \armman{}, we demonstrate significant deviations from the Markov assumption. To solve the challenges that arise, we model arms as time-series, and propose the \fullname{} (\proposed{}) policy, a novel algorithm that selects the arms that will benefit the most from an intervention, given our future state predictions. Our evaluation shows a significant increase in engagement compared to the SOTA, deployed Whittle index solution, on real data from \armman{} with \numparticipats{} participants. This translates to $16.3$ hours of additional content listened, $90.8\%$ more engagement drops prevented, and reaching more than twice ($\times2.35$) as many high dropout-risk beneficiaries. While we focus on maternal and child healthcare as an indicative application, we expect the proposed approach to perform well on any other RMAB application that involves non-Markovian behaviours.

\section*{Acknowledgments} \label{sec: Acknowledgments}

P.D. was funded in part by JPMorgan Chase \& Co. Any views or opinions expressed herein are solely those of the authors listed, and may differ from the views and opinions expressed by JPMorgan Chase \& Co. or its affiliates. This material is not a product of the Research Department of J.P. Morgan Securities LLC. This material should not be construed as an individual recommendation for any particular client and is not intended as a recommendation of particular securities, financial instruments or strategies for a particular client. This material does not constitute a solicitation or offer in any jurisdiction.

J.A.K. was supported by an NSF Graduate Research Fellowship under grant DGE1745303


\small
\bibliographystyle{named}
\bibliography{ijcai23_full}

\appendix

\section*{Contents}

In this appendix we include several details that have been omitted from the main text due to space limitations. In particular:

\begin{itemize}
    \item[-] In Section \ref{supp: TSF Model Architecture and Implementation Details}, we describe and evaluate various time series forecasting (TSF) models.
    \item[-] In Section \ref{supp: Whittle Index Policy}, we provide the definition of the Whittle index policy.
    \item[-] In Section \ref{supp: Estimating Transition Probabilities for Whittle Policy}, we describe how we calculate the empirical transition probabilities of the underlying Markov model for the Whittle index baseline.
    \item[-] In Section \ref{supp: Synthetic Data}, we explain how we generate the synthetic agents.
    \item[-] In Section \ref{supp: Counterfactuals in Non-Markovian Settings Alternative}, we provide an alternative approach to computing counterfactuals in a non-Markov setting.
    \item[-] In Section \ref{supp: Real-World Dataset}, we provide additional details on the real data on maternal and child healthcare awareness from our partner NGO, \armman{}. 
    \item[-] In Section \ref{supp: Ethics, Consent, Data Use and Information Accessibility}, we comment on the ethics and consent of data usage.
    \item[-] In Section \ref{supp: Limitations}, we describe the limitations of the proposed approach.
    \item[-] In Section \ref{supp: Language and Terminology}, we provide a short note on the language used throughout the paper.
    \item[-] In Section \ref{supp: High Dropout-Risk Beneficiaries}, we give the definition for the high-dropout risk beneficiaries (`critical' beneficiaries).
    \item[-] Finally, Sections \ref{supp: Detailed Evaluation Results: Synthetic Data}, and \ref{supp: Detailed Evaluation Results: Real Data}, provide detailed numerical results for all our simulations on both synthetic and real data.
\end{itemize}

\section{TSF Model Architecture and Implementation Details} \label{supp: TSF Model Architecture and Implementation Details}

We implemented several architectures for TSF, of varying complexity. Specifically:

\begin{enumerate}
    \item Long short-term memory (LSTM)~\cite{hochreiter1997long}: Capable of learning long-term dependencies by preserving information from inputs in hidden states.
    \item Bidirectional LSTM (BiLSTM)~\cite{graves2005framewise}. The input is processed in two ways, one from past to future and one from future to past.
    \item BiLSTM with an Attention Layer~\cite{vaswani2017attention} (AttentionBiLSTM): Adding an attention layer can help a model with large sequences of data by attending (focusing on) important parts of the input, while diminishing others.
    \item  Sequence to Sequence Model (Seq2SeqAttention)~\cite{luong2015effective}: Encoder-decoder architecture.
    \item BiLSTM with timeseries embeddings (Time2VecBiLSTM)~\cite{kazemi2019time2vec}: The Time2Vec embedding is invariant to time re-scaling and can capture periodic and non-periodic patterns.
    \item Transformer with timeseries embedding (Time2VecTransformer): Time2Vec embeddings with a transformer~\cite{vaswani2017attention} block.
    \item Time series Transformer (TSTransformer): A transformer architecture for multivariate time series forecasting.
\end{enumerate}
 
We perform min-max scaling on all the data before training. We used the Adam optimizer~\cite{kingma2014adam}, with learning rate $1e-4$, and minibatch size $64$. To evaluate our models, we performed a walk-forward validation via recursive multi-step forecasting -- i.e., attempted to re-create the test trajectories using model forecasts -- and measured the mean absolute error (MAE) over all trajectories and timesteps. Tables \ref{tab: synthetic data model error}, and \ref{tab: real data model error} depict the results for our various models, for synthetic, and real data trajectories, respectively. We report results on single step prediction, and full trajectory prediction. The latter corresponds to 52 timesteps for synthetic data (1 year), and 23 timesteps for the real data ($>5$ months). Most of our models perform well, especially considering the difficulty of the forecasting task under long time horizons.

\begin{table}[t!]
\centering
\begin{tabular}{@{}rcc@{}}
\toprule
MODEL               & MAE @ 1step & MAE @ end \\ \midrule
LSTM                & 0.034262    & 0.228129  \\
BiLSTM              & 0.035208    & 0.147303  \\
AttentionBiLSTM     & 0.041949    & 0.193646  \\
Seq2SeqAttention    & 0.109652    & 0.355122  \\
Time2VecBiLSTM      & 0.025425    & 0.150013  \\
Time2VecTransformer & 0.057951    & 0.220191  \\
TSTransformer       & 0.054733    & 0.219718  \\ \bottomrule
\end{tabular}%
\caption{Mean absolute error (MAE) for 1 step and 52 steps (entire trajectory) ahead prediction, for various TSF models, on our synthetic data.}
\label{tab: synthetic data model error}
\end{table}

\begin{table}[t!]
\centering
\begin{tabular}{@{}rcc@{}}
\toprule
MODEL                & MAE@1step & MAE@end  \\ \midrule
LSTM                 & 0.200043  & 0.381700 \\
BiLSTM               & 0.202291  & 0.376472 \\
AttentionBiLSTM      & 0.200991  & 0.346825 \\
Seq2SeqAttention     & 0.211980  & 0.398529 \\
Time2VecBiLSTM       & 0.195265  & 0.342587 \\
Time2VecTransformer  & 0.201717  & 0.371806 \\
TSTransformer        & 0.213810  & 0.402264 \\ \bottomrule
\end{tabular}
\caption{Mean absolute error (MAE) for 1 step and 23 steps (entire trajectory) ahead prediction, for various TSF models, on real data on maternal and child healthcare awareness from \armman{}.}
\label{tab: real data model error}
\end{table}

\section{Whittle Index Policy} \label{supp: Whittle Index Policy}

Informally, the Whittle index of an arm captures the added value from pulling said arm. Consider a `passive subsidy' -- a hypothetical exogenous compensation $m$ rewarded for not pulling ($a=0$) an arm $i$. The Whittle index is defined as the smallest subsidy necessary to make the planner indifferent between pulling and not pulling (assuming indexability~\cite{liu2010indexability}), specifically:

\begin{definition}[Whittle index]\label{def:whittle-index}
The Whittle index associated to state $s \in \mathcal{S}$ of arm $i$ is defined by:
\begin{align}
    W_i(s) & \triangleq \inf\nolimits_{m}
    \{ Q_{i}^m(s \mid a=0) = Q_{i}^m(s \mid a=1)\} \label{eqn:action-indifference}
\end{align}
where $Q_{i}^m(\cdot)$ is given by the following augmented Bellman equations:
\begin{align*}
    Q_{i}^m(s \mid a) &= m \mathds{1}_{a = 0} + R(s) + \gamma \sum_{s'} \Pr\nolimits_{i}(s,a,s') V^{m}_{i}(s') \\
    V^{m}_i(s) &=  \max_{a} Q_{i}^m(s \mid a)
\end{align*}
\end{definition}

The Whittle index policy computes the $W_i(s)$ of all arms and pulls the arms with the highest values of the index at each timestep. The Bellman equations are solved via value iteration, and binary search is used to find the smallest $m$ that satisfies Equation~\ref{eqn:action-indifference}.

We used the two-stage Whittle index as a baseline (e.g., see~\cite{mate2022field}).

\section{Estimating Transition Probabilities for Whittle Policy} \label{supp: Estimating Transition Probabilities for Whittle Policy}

To compute the Whittle indices, we must know beforehand the transition probabilities parametrizing the Markov Decision Process. This is not always true in the real world. Similar to \cite{mate2022field}, we learn a model which predicts a beneficiarie's transition probabilities given their demographic features. Specifically:
\begin{enumerate}
    \item Based on historical data, we estimate beneficiaries' empirical transition probabilities.
    \item Since many beneficiaries would have missing transitions, we cluster beneficiaries in the feature space and calculate empirical transition probabilities for a given cluster of beneficiaries.
    \item Given beneficiaries' demographic features, we learn a mapping from features to behaviour cluster.
    \item For beneficiaries in the test set, we use the mapping model to predict which cluster a beneficiary belongs to and thus obtain their estimated transition probability.
\end{enumerate}
\section{Synthetic Data}  \label{supp: Synthetic Data}

We generate a dataset of trajectories containing an equal number of the following agents:

\paragraph{Habit former} Starts at state $s = 0.75 \pm \mathcal{U}(0.2)$, where $\mathcal{U}(x)$ denotes uniform noise in $[0, x]$. State drops under $a=0$ with rate $0.03 \pm \mathcal{U}(0.03)$ (minimum drop rate of 0.01). State increases under $a = 1$, as follows: $s' = (1 + (0.2 + \mathcal{U}(0.2))) \times s$. The aforementioned noise levels are drawn once at the creation of each agent, and remain constant during the trajectory. If the state reaches 1 (formed a habit), it stays there -- independent of the action -- for a duration drawn uniformly at random from [8, 12].

For the testing dataset, the drop rate is set to $0.1 \pm \mathcal{U}(0.05)$, while the increase rate is set to $0.2 + \mathcal{U}(0.1)$.

\paragraph{Motivation based} Starts at state $s = 1 - \mathcal{U}(0.2)$. State drops over time under $a=0$ at a rate of $0.05 \pm \mathcal{U}(0.05)$ (minimum drop rate of 0.01). If we act, the state returns to baseline.

For the testing dataset, the drop rate is set to $0.1 \pm \mathcal{U}(0.05)$.

\paragraph{Random} Random state independent of the action.

\section{Counterfactuals in Non-Markovian Settings -- Alternative Approach} \label{supp: Counterfactuals in Non-Markovian Settings Alternative}

An alternative approach to computing counterfactuals in non-Markov environments would be to use our TSF model to compute the counterfactual for only a single step when trajectories deviate (in order to calculate efficiency gains), and then remove the deviated participants from the simulation (i.e., arms would decrease over time). The advantage of this approach is that we avoid the potential error accumulation on the counterfactual trajectories over time. The drawback is that we do not account for long terms gains from acting. We have evaluated both approaches, with \proposed{} consistently outperforming the baselines in both cases. See Section \ref{supp: Detailed Evaluation Results: Real Data} for detailed results.

\section{Real-World \armman{} Dataset} \label{supp: Real-World Dataset}

We use historical data from a large-scale quality improvement study performed by \armman{} in 2022, obtained with beneficiary consent. The data follows 12000 participants (11256 with complete state information by the end) over a period of 31 weeks. In the study, a set of beneficiaries received interventions from a variety of policies such as the current standard of care, where there are no service calls, a Random Policy, and a Whittle Index based policy. Each beneficiary is represented by a single trajectory of states (engagement behavior) and actions (received, or not a call from a healthworker). Additionally, we have 43 static features for every beneficiary collected at the time of registration. These describe beneficiaries' characteristics such as age, gestational age at time of registration, family income, education level, etc.

\subsection{Feature List}
We provide the full list of 43 features used for predicting the transition probabilities for the Whittle index baseline (see Section \ref{supp: Estimating Transition Probabilities for Whittle Policy}):
\begin{itemize}
    \item Enrollment gestation age
    \item Age (split into 5 categories)
    \item Family income (8 categories)
    \item Education level (7 categories)
    \item Language (5 categories)
    \item Phone ownership (3 categories)
    \item Call slot preference (5 categories)
    \item Enrollment channel (3 categories)
    \item Stage of pregnancy
    \item Days since first call
    \item Gravidity, parity, stillbirths, live births
\end{itemize}

\section{Ethics, Consent, Data Use, and Information Accessibility} \label{supp: Ethics, Consent, Data Use and Information Accessibility}

\subsection{Ethics}

We understand the responsibility associated with AI-systems for undeserved communities. Prior to any experimentation, ethics approval was obtained from the NGO's Ethics Review Board. Additionally, experts from the NGO were kept in the loop regarding experimentation and model development process.

\subsection{Secondary Analysis and Data Usage}

Our work is categorized as secondary analysis of the data described in previous section. The models are trained using data collected from past engagement behaviors of beneficiaries in the mobile health program. 

None of the proposed approaches involve deployment. Necessary approvals were obtained from the ethics review board of the NGO prior to performing the secondary data analysis.

\subsection{Consent and Data Sharing}

Consent is received from the beneficiaries for  participating in the program. All the data collected through the program is owned by the NGO and only the NGO is allowed to share the data. 
The data have read-only access for researchers and are shared by the NGO through clearly defined exchange protocols after approval by NGO's ethics review committee.

\subsection{Universal Accessibility of Health Information}

The service call scheduling algorithm only optimizes the quality of service calls and does not withhold health information from beneficiaries. Irrespective of whether a beneficiary receives a service call or not (intervention), they still get weekly automated voice messages on health information. Additionally, beneficiaries can request a service call themselves through a free missed call service.

\section{Limitations} \label{supp: Limitations} 

This work provides a major step forward in RMAB research, opening avenues to previously understudies areas like non-Markov environments, and continuous states. Yet, the proposed approach relies on the existence of a reasonably well performing time-series forecasting (TSF) model. Our evaluation demonstrates that state-of-the-art TSF models do achieve high performance in both synthetic, and real data, and long time horizons. Yet, this might not be the case for every domain. Nevertheless, our approach is model agnostic, thus future advancements in the area of time series forecasting can easily translate to better performance for the proposed approach. Finally, \proposed{} was designed taking into account domain characteristics, notably the limited resources. As such, the performance of \proposed{} remains an open question in a domain where each arm receives a large number of interventions (large budget).

\section{Language and Terminology} \label{supp: Language and Terminology} 

In this paper, we use the term mother to refer to pregnant, birthing, and postnatal people. We recognize that the term may not reflect the identity of all people, and we stand by the need to be inclusive. While the term might not be perfect -- in lack of a better, well established, and accepted alternative -- we opted to use it to keep our writing concise.

\begin{table}[t]
\centering
\begin{tabular}{@{}rcc@{}}
\toprule
Algorithm       & Mean       & Max         \\ \midrule
vs. Whittle     & 44.2 (\%)  & 107.3 (\%)  \\
vs. Round-robin & 102.5 (\%) & 174.6 (\%)  \\
vs. Random      & 128.6 (\%) & 186.6 (\%)  \\ \bottomrule
\end{tabular}%
\caption{Synthetic data, $N = 90$ arms, budget $k = 9$. Relative improvement (\proposed{} vs. each algorithm) in engaging beneficiaries (excluding the random agents). Control is missing as all beneficiaries eventually become non-engaging.}
\label{tab: synthetic algos}
\end{table}

\section{High Dropout-Risk Beneficiaries (`Critical' Beneficiaries)} \label{supp: High Dropout-Risk Beneficiaries}

We label as `critical' a beneficiary who (i) had at least one engaging call in the first 6 weeks, (ii) never received a service call in the ground truth trajectories, and (iii) stopped engaging with the calls continuously for 6 weeks at the end of our observation period.

\section{Detailed Evaluation Results: Synthetic Data} \label{supp: Detailed Evaluation Results: Synthetic Data}

In this section, we provide detailed numerical results on our simulations on synthetic data. Specifically, Tables \ref{tab: synthetic algos}, \ref{tab: synthetic arms}, \ref{tab: synthetic budget}, \ref{tab: synthetic bins}, and \ref{tab: synthetic history}, present results on comparison with different baselines, varying the number of arms ($N$), varying the intervention budget ($k$), having a finer discretization of the state for the Whittle index baseline, and adding history to the Whittle index baseline, respectively. Standard deviation values on the percentage of engagement are up to $3.9\%$, $326.6\%$, $431.1\%$, and $0\%$, for \proposed{}, Whittle index, Round-robin, and control, respectively.

\begin{table}[t]
\centering
\begin{tabular}{@{}rcc@{}}
\toprule
Arms $N$ & Mean     & Max      \\ \midrule
30       & 48.2 (\%) & 158.7 (\%) \\
90       & 44.2 (\%)  & 107.3 (\%) \\
120      & 47.3 (\%) & 141.8 (\%) \\
600      & 48.2 (\%) & 114.9 (\%) \\ \bottomrule
\end{tabular}%
\caption{Synthetic data. Relative improvement  of \proposed{} (compared to Whittle index, the best baseline) in engaging beneficiaries (excluding the random agents). Varying number of arms $N$, budget $k = 0.1 \times N$.}
\label{tab: synthetic arms}
\end{table}

\begin{table}[t]
\centering
\begin{tabular}{@{}rcc@{}}
\toprule
Budget $k$ & Mean     & Max      \\ \midrule
1          & 37.8 (\%) & 94.6 (\%) \\
9          & 44.2 (\%)  & 107.3 (\%) \\
18         & 88.7 (\%) & 221.3 (\%) \\ \bottomrule
\end{tabular}%
\caption{Synthetic data. Relative improvement  of \proposed{} (compared to Whittle index, the best baseline) in engaging beneficiaries (excluding the random agents). Varying budget $k$. Number of arms $N = 90$.}
\label{tab: synthetic budget}
\end{table}

\begin{table}[t]
\centering
\begin{tabular}{@{}rcc@{}}
\toprule
Bins & Mean     & Max      \\ \midrule
2    & 44.2 (\%)  & 107.3 (\%) \\
5    & 114.8 (\%) & 248.8 (\%) \\
9    & 138.1 (\%) & 345.8 (\%) \\ \bottomrule
\end{tabular}%
\caption{Synthetic data, $N = 90$ arms, budget $k = 9$. Relative improvement  of \proposed{} (compared to Whittle index, the best baseline) in engaging beneficiaries (excluding the random agents). Varying number of bins for discretizing the state for the Whittle index baseline.}
\label{tab: synthetic bins}
\end{table}

\begin{table}[t]
\centering
\begin{tabular}{@{}rcc@{}}
\toprule
History   $h$ & mean     & max      \\ \midrule
1             & 44.2 (\%)  & 107.3 (\%) \\
2             & 67.0 (\%) & 322.7 (\%)  \\
3             & 69.6 (\%) & 145.7 (\%) \\
5             & 144.2 (\%) & 467.3 (\%) \\ \bottomrule
\end{tabular}%
\caption{Synthetic data, $N = 90$ arms, budget $k = 9$. Relative improvement  of \proposed{} (compared to Whittle index, the best baseline) in engaging beneficiaries (excluding the random agents). Varying history given (in the form of expanded state $s' = \times_{i=1}^h s_i$) to the Whittle index baseline.}
\label{tab: synthetic history}
\end{table}

\section{Detailed Evaluation Results: Real Data on Maternal and Child Healthcare Awareness} \label{supp: Detailed Evaluation Results: Real Data}

\begin{figure}[th!]
    \centering
    \includegraphics[width=\linewidth, trim={0em 0.8em 0em 0.7em}, clip]{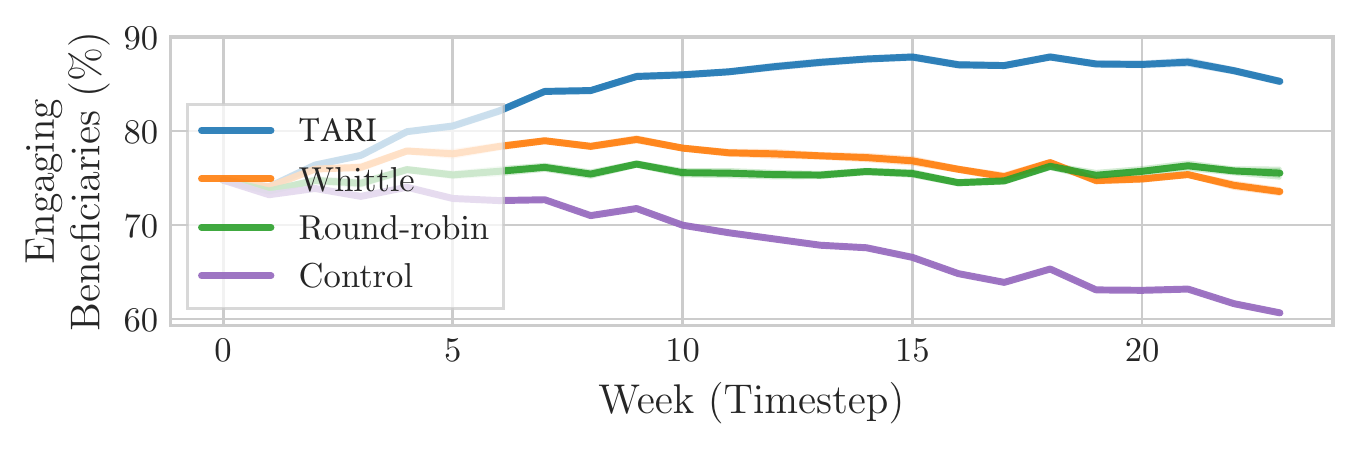}
    \caption{Percentage of engaging beneficiaries. \proposed{} achieves up to $17.7\%$ higher engagement compared to Whittle. Note that the standard deviation is small, as the only source of randomness is tie-breaking arms (there are no different datasets as with synthetic data).}
    \label{fig: results real data}
\end{figure}

Figure \ref{fig: results real data} shows the percentage of engaging beneficiaries during the time horizon. The proposed approach, \proposed{}, achieves the highest engagement, \emph{up to $17.7\%$} improvement relative to the Whittle index. Moreover this figure suggests that the \proposed{} would benefit from a larger time horizon, as the gap increases over time (our evaluation was limited by the length of the real trajectories). Similar results were achieved for budgets of $3\%$, and $4\%$, and for the second evaluation approach where we remove arms that deviate from historical trajectories (see Tables below).

To measure the real-world impact of our proposed model in comparison with baseline policies, we use 6 metrics as described below:

\begin{itemize}
    \item Mean Weekly Engagement Improvement: This metric measures how many more beneficiaries are engaging on average in the \proposed{} policy as compared to the baseline policy.
    \item Mean Relative Engagement Improvement: This metric measures the mean relative improvement in number of beneficiaries engaging every week in the \proposed{} policy as compared to baseline policy.
    \item Cumulative Additional Engagement: This metric cumulatively measures the increase in beneficiary engagement through \proposed{} as compared to baseline policy.
    \item Cumulative Additional Duration: This metric cumulatively measures the increase in duration of calls listened to (in seconds) through \proposed{} as compared to baseline policy.
    \item Relative Increase in Cumulative Additional Engagement: This metric measures the relative increase in Cumulative Additional Engagement in \proposed{} as compared to baseline policy.
\end{itemize}

In Tables \ref{tab: real data 1} - \ref{tab: real data 4}, we report these metrics for various settings such as increasing budget (2\%, 3\%, and 4\%), and for two evaluation approaches described in Section \ref{sec: Counterfactuals in Non-Markovian Settings}, and Section \ref{supp: Counterfactuals in Non-Markovian Settings Alternative}. In all the settings, \proposed{} achieves strong gains over all the baselines.

\begin{table*}[t!]
\centering
\begin{tabular}{rcccc}
\hline
\textbf{}                                                & \textbf{Whittle} & \textbf{Round-robin} & \textbf{Control} \\ \hline
\textbf{Mean weekly engagement improvement}              & 175.230435       & 202.726087          & 362.521739       \\
\textbf{Mean relative engagement improvement}            & 10.200858        & 11.919693           & 24.322864        \\
\textbf{Cumulative additional engagement}                & 4030.300000      & 4662.700000         & 8338.000000      \\
\textbf{Cumulative additional duration}                  & 58969.050550     & 170972.589044       & 225619.226922    \\
\textbf{Relative inc. cumulative additional engagement}  & 10.159053        & 11.935361           & 23.576316        \\ \hline
\end{tabular}
\caption{Improvement through \proposed{} over other baselines as measured through different metrics. Evaluation Method 1, Budget 2\%}
\label{tab: real data 1}
\end{table*}

\begin{table*}[t!]
\centering
\begin{tabular}{rcccc}
\hline
\textbf{}                                                & \textbf{Whittle} & \textbf{Round-robin} & \textbf{Control} \\ \hline
\textbf{Mean weekly engagement improvement}              & 194.952174       & 184.600000          & 424.004348       \\
\textbf{Mean relative engagement improvement}            & 11.013070        & 10.384419           & 28.247768        \\
\textbf{Cumulative additional engagement}                & 4483.900000      & 4245.800000         & 9752.100000      \\
\textbf{Cumulative additional duration}                  & 51261.623810     & 198894.898049       & 280280.976819    \\
\textbf{Relative inc. cumulative additional engagement}  & 11.034964        & 10.388184           & 27.574789        \\ \hline
\end{tabular}
\caption{Improvement through \proposed{} over other baselines as measured through different metrics. Evaluation Method 1, Budget 3\%}
\label{tab: real data 2}
\end{table*}

\begin{table*}[t!]
\centering
\begin{tabular}{rcccc}
\hline
\textbf{}                                                & \textbf{Whittle} & \textbf{Round-robin} & \textbf{Control} \\ \hline
\textbf{Mean weekly engagement improvement}              & 189.213043       & 135.121739          & 458.100000       \\
\textbf{Mean relative engagement improvement}            & 10.427997        & 7.463115            & 30.428595        \\
\textbf{Cumulative additional engagement}                & 4351.900000      & 3107.800000         & 10536.300000     \\
\textbf{Cumulative additional duration}                  & 41746.542438     & 217412.110742       & 328153.881461    \\
\textbf{Relative inc. cumulative additional engagement}  & 10.473867        & 7.262442            & 29.792173        \\ \hline
\end{tabular}
\caption{Improvement through \proposed{} over other baselines as measured through different metrics. Evaluation Method 1, Budget 4\%}
\label{tab: real data 3}
\end{table*}

\begin{table*}[t!]
\centering
\begin{tabular}{rcccc}
\hline
\textbf{}                                                & \textbf{Whittle} & \textbf{Round-robin} & \textbf{Control} \\ \hline
\textbf{Mean weekly engagement improvement}              & 210.640457       & 337.889670          & 348.913858       \\
\textbf{Mean relative engagement improvement}            & 12.841306        & 22.598831           & 23.496397        \\
\textbf{Cumulative additional engagement}                & 4844.730509      & 7771.462417         & 8025.018744      \\
\textbf{Cumulative additional duration}                  & 72610.090354     & 163440.195309       & 217932.682347    \\
\textbf{Relative inc. cumulative additional engagement}  & 12.568732        & 21.819042           & 22.691338        \\ \hline
\end{tabular}
\caption{Improvement through \proposed{} over other baselines as measured through different metrics. Evaluation Method 2 (Section \ref{supp: Counterfactuals in Non-Markovian Settings Alternative}), Budget 2\%}
\label{tab: real data 4}
\end{table*}

\end{document}